\documentclass[sigconf,natbib=false]{acmart}

\AtBeginDocument{%
  }

\RequirePackage[
  datamodel=acmdatamodel,
  style=acmnumeric,sorting=none
  ]{biblatex}
\copyrightyear{2024}
\acmYear{2024}
\setcopyright{acmlicensed}\acmConference[WWW '24 Companion]{Companion Proceedings of the ACM Web Conference 2024}{May 13--17, 2024}{Singapore, Singapore}
\acmBooktitle{Companion Proceedings of the ACM Web Conference 2024 (WWW '24 Companion), May 13--17, 2024, Singapore, Singapore}
\acmDOI{10.1145/3589335.3651905}
\acmISBN{979-8-4007-0172-6/24/05}

\begin{document}

\title{Dynamic Contexts for Generating Suggestion Questions in RAG Based Conversational Systems}

\author{Anuja Tayal}
\authornote{work done as part of internship at the Procter and Gamble Company.}
\orcid{0009-0002-3631-8195}
\affiliation{%
  \institution{University of Illinois, Chicago}
  \department{Department of Computer Science}
  \city{Chicago}
  \state{IL}
  \country{USA}
}
\email{atayal4@uic.edu}

\author{Aman Tyagi}
\affiliation{%
  \institution{The Procter and Gamble Company}
  \city{Mason}
  \state{Ohio}
  \country{USA}
}
\email{tyagi.a.3@pg.com}


\begin{abstract}

When interacting with Retrieval-Augmented Generation (RAG)-based conversational agents, the users must carefully craft their queries to be understood correctly. Yet, understanding the system's capabilities can be challenging for the users, leading to ambiguous questions that necessitate further clarification. This work aims to bridge the gap by developing a suggestion question generator. To generate suggestion questions, our approach involves utilizing dynamic context, which includes both dynamic few-shot examples and dynamically retrieved contexts. Through experiments, we show that the dynamic contexts approach can generate better suggestion questions as compared to other prompting approaches.

\end{abstract}

%
%





\maketitle
\section{Introduction}
Current conversational systems encounter numerous problems. Firstly, they depend heavily on user inputs, with the agent merely responding to queries \cite{quac,coqa}. This method is fundamentally flawed as it fails to actively prompt users to clarify ambiguous or poorly defined questions. As a result, the systems frequently return incorrect responses or resort to apologetic statements when dealing with such inquiries. Secondly, even when the system does provide a correct answer, the user must again formulate an effective follow-up question. This task can be challenging, as users often find it difficult to understand the full scope of the system's capabilities \cite{tod-cq-clarit}. This cycle of inadequate responses forces users to constantly refine their queries, often resulting in more apologies than useful information.

Through this paper, we want to address this change and fill the gap by proposing a suggestion question generator. The generator is designed to generate suggestion questions that the agent can answer, guided by the user's initial query. Our framework encompasses several key steps. First, it involves identifying pertinent passages (dynamically retrieved contexts) based on the user's initial query, as the answer to the query hinges on these passages. Secondly, we take the help of relevant triple of (questions, answer, suggestion questions) (dynamic few-shot). Finally, we generate suggestion questions by constituting a dynamic context prompt consisting of the dynamically retrieved contexts, and dynamic few-shot examples. By providing suggestion questions, users are alleviated from the task of question formulation, ensuring a smooth conversational flow \cite{tod-cq-clarit}. 

Conversational agents can adapt Dynamic Contexts for a variety of tasks, eliminating the need for extensive model training. To generate suggestion questions in the absence of a publicly available dataset, we use in-context learning. In-context learning \cite{few-shot} has been recognized for achieving promising outcomes with minimal training samples. 

\begin{figure}[h]   
    \includegraphics[width=\linewidth]{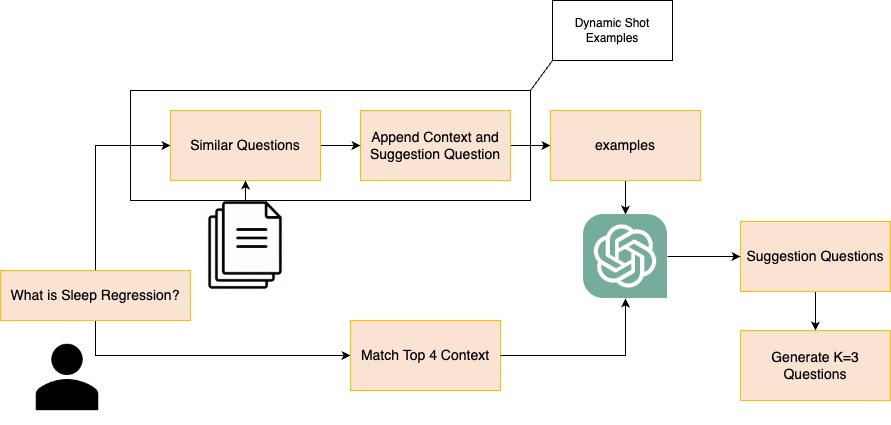}
    \caption{Dynamic Context: Suggestion Questions are generated in RAG based Chatbots using the initial user query along with dynamically few-shot examples and dynamic retrieved contexts}
    \label{fig:dynamic-shot}
    \vspace{-10pt}
\end{figure}

\section{Problem Statement}
Our primary goal is to develop a suggestion question generator. We aim to tackle cases where a RAG-based system cannot interpret a user's query accurately, despite the presence of relevant information in the dataset. Due to this, the system is unable to understand the query's intent leading to unsuccessful responses. Hence, we generate three suggested questions that the agent is capable of responding based on the user's initial query. 

In Summary, in this paper, we:
\begin{itemize}
    \item Propose to develop a suggestion question generator to address the gap in conversational systems where users frequently lack awareness of the system's limitations.
    \item Formulate a suggestion question generator in low-resource settings which can be applied across various datasets without the need of fine-tuning.
    \item Propose a dynamic context prompt that constitutes dynamic few-shot examples and dynamic retrieved contexts to generate suggestion questions.
\end{itemize}
\section{Related Work}
To the best of our knowledge, there is no existing model or literature specifically addressing suggestion question generation. However, this concept is closely related to question generation \cite{sparta} and clarification question generation \cite{tod-cq-clarit,claqua}. 

The concept of clarification questions was formally introduced in \cite{cq1}, and since then, research into generating these questions has spanned a wide range of scenarios, including open-domain systems (AmbigQA) \cite{ambigqa}, knowledge bases (CLAQUA) \cite{claqua}, closed-book systems (CLAM) \cite{clam}, information-seeking (ISEEQ) \cite{iseeq}, task-oriented dialog systems (CLARIT) \cite{tod-cq-clarit}, and conversational search \cite{mixed-initiative}. Rahmani et al. \cite{rahmani-etal-2023-survey} surveyed the various methodologies, datasets, and different evaluation strategies used for clarification questions.

The distinct characteristics of RAGs \cite{dpr} make them essential for effective information retrieval and is widely used for generating question-answering tasks \cite{quac} and clarification questions \cite{iseeq, mixed-initiative}.

Advancements in large language models (LLMs) have led researchers to investigate the use of LLMs and prompting techniques for both question generation \cite{sparta} and clarification questions \cite{deng2023prompting}.

\section{Dynamic Contexts Approach}
Dynamic Contexts, the Suggestion Question Generator (Figure \ref{fig:dynamic-shot}), is designed to enhance the interaction between users and conversational systems by addressing the gap in users' understanding of these systems' capabilities. By aiming to reduce the frequency of apologetic responses \cite{tod-cq-clarit}, it provides more informed system suggestions, thereby improving the overall user experience.

Dynamic Contexts generate suggestion questions that are answerable by the agent. This is achieved through two principal components of Dynamic Contexts: dynamic few-shot examples and dynamically retrieved contexts. Dynamic few-shot examples begin with the selection of contextually relevant triplets from the dataset, each consisting of a question, its answer, and the suggestion question (QAS). Dynamic few-shot diverges from traditional few-shot prompting by dynamically choosing each example based on the user's query, rather than relying on a static set of examples. This dynamic selection is crucial for accommodating the varied formats and structures that different questions may necessitate. 
 
To ensure that the questions generated by the Suggestion Question Generator remain pertinent and answerable, dynamic contexts consist of dynamically retrieved contexts.  Dynamically retrieved contexts consist of retrieving 4 dynamic contexts based on the initial user query, similar to RAG-based systems from which suggestion questions should be generated. We use OpenAI embeddings to embed the examples and dynamic contexts and cosine similarity to select dynamic examples and contexts. The complete prompt structure utilized for the approach is detailed in the appendix \ref{sec:appendix}.


\begin{table} [t]
\scriptsize
     \begin{tabular}{|p{.195\textwidth}|p{.075\textwidth}|p{.075\textwidth}|p{.05\textwidth}|}
    \hline
            & \textbf{ChatGPT} & \textbf{Claude-2} & \textbf{GPT-4} \\
         \hline
         Zero-Shot &35 &30 & 43 \\
         \hline
         Few-Shot & 42& 35 & 40 \\
         \hline
         Dynamic Few-Shot & 42 & 35 &43\\
         \hline
         Dynamic Contexts (Our approach) & 44 & 44 &46\\
    \hline
    \end{tabular}
    \caption{Comparative Analysis of Dynamic Contexts with Zero-Shot, Few-Shot and Dynamic Few-Shot}
    \label{table:Analysis}
    \vspace{-20pt}
\end{table}  

\section{Evaluation}
Due to the absence of a publicly accessible dataset specifically designed for Suggestion Question Generator tasks, we employed 228 blog posts from Pamper's Baby Sleep Coach as a practical demonstration of the approach. More details of the dataset are present in Appendix \ref{dataset}. To test the Dynamic Contexts approach, LLMs- ChatGPT, GPT-4, and Claude2 were used. The evaluation of the suggestion question generator involves a variety of assessment techniques to ensure a thorough analysis of the performance. 
 
\textbf{Manual Evaluation} To assess the effectiveness of the Dynamic Contexts approach in the Suggestion Question Generator, we conducted a manual evaluation using a set of 48 QAS (question, answer, and suggestion question) pairs. During manual assessment, we primarily assessed the correctness, relevance, and soundness of the generated questions. Dynamic Contexts demonstrated its capability by successfully generating relevant suggestion questions (as shown in Table \ref{table:Analysis}). ChatGPT, Claude-2, and GPT-4 each produced 44, 44, and 46 correct samples, respectively. ChatGPT and GPT-4 errors were related to the baby's age, whereas Claude-2 did not make any age-related errors. However, Claude-2's primary shortcomings were its failure to formulate questions that aligned with the provided contexts.

\textbf{Comparative Analysis} A comparative analysis was performed to evaluate the improvement of Dynamic Contexts over different prompting strategies, including Zero-Shot, Few-Shot, and Dynamic Few-Shot approaches. This comparison, detailed in Table \ref{table:Analysis}, highlights the distinctions among the methods. The Zero-Shot method employed no examples in its prompts, while the Few-Shot method used a consistent set of three examples for each query. The Dynamic Few-Shot approach, in contrast, dynamically selected its examples. Distinct from other methods, Dynamic Contexts utilizes Retrieval-Augmented Generation (RAG) to dynamically select contexts for generating suggestion questions. This approach marks a significant shift from the static contexts used in the Zero-Shot, Few-Shot, and Dynamic Few-Shot approaches. 

The findings indicate that Claude-2 generated only 30 correct questions (compared to 35 for ChatGPT and 43 for GPT-4) in the zero-shot approach out of the 48 samples. While we supplied the answers, the models struggled to generate questions that could be answered from the provided context. Claude-2 notably struggled, often unable to utilize the provided context effectively, and sometimes misconstrued its role, imagining itself as a baby.

The few-shot approach, which involved providing three example questions to the models, showed an overall improvement in performance. However, the questions generated tended to replicate the structure of the examples provided too closely. The dynamic few-shot approach, aimed at introducing a variety of question types and structures, significantly improved the quality of the questions generated but still lagged behind the dynamic contexts approach. 

\textbf{Preference Benchmarking} Further analysis was conducted through Preference Benchmarking Evaluation, where GPT-4 and Claude-2 evaluated the suggestion questions generated by the framework. In this blind test, the models, GPT-4 and Claude-2 acted as judges and were prompted to assess the quality of the questions without knowing from which model it was generated. The results showed Claude2 having no preference at all between the models, while GPT-4 favored Claude-2's output 57\% of the time over its own. Human evaluation was also employed to evaluate which output was better which showed a preference for GPT-4's output in 43\% of cases, Claude-2 in 33\% of cases, and showing no preference in 24\% of cases. This evaluation suggests a close alignment between the judgments of LLMs and human evaluators. Detailed results can be found in Table \ref{table:Preference-BenchMarking} of the appendix \ref{sec:benchmark}.

\textbf{Ablation Study} To determine the optimal order of the final prompt, we conducted an ablation study, altering the arrangement of example contexts and queries. Initially, we presumed that presenting the query before the dynamic contexts would enhance comprehension, allowing the model to suggest questions based on the given information. However, upon changing the order, we did not observe any discernible difference in the nature of the generated suggestion questions as seen from the Table \ref{table:Ablation-Study} in Appendix \ref{ablation}. 

\section{Conclusion}
In this paper, we explore suggestion questions to bridge the gap in users' comprehension of a conversational system's capabilities. We present the development of a suggestion question generator that utilizes dynamic context prompting within a RAG-based conversational system. In the future, we plan to personalize the suggestion questions based on the user history.

\begin{acks}
The authors would like to acknowledge the support of the Procter and Gamble Company (P\&G). Any opinions, findings, and conclusions or recommendations expressed in this material are those of the authors and do not necessarily reflect the views of P\&G.
\end{acks}

\printbibliography

\appendix
\section{Dataset}
\label{dataset}
Due to the absence of a publicly accessible dataset specifically designed for Suggestion Question Generator tasks, we employed 228 blog posts from Pamper's Baby Sleep Coach as a practical demonstration of the approach. The blog post headlines were treated as queries, while the content of the posts as the answers to those queries. For 35 of these posts, we manually annotated suggestion questions aiming to exemplify the types of questions users might ask, and which could be answered by the provided context. The post headlines from these 35 examples form the basis for identifying queries similar to the initial user query. To each of these similar queries, we then append the corresponding answer and a suggestion question, forming a set of dynamic few-shot examples consisting of pairs of a similar query, its answer, and a suggestion question (as shown in Figure \ref{fig:dynamic-shot}).  Meanwhile, the original set of 138 answers forms the basis for our answer contexts, from which we select the top 4 contexts. The complete prompt structure utilized for the approach is detailed in the appendix \ref{sec:appendix}.

The dataset presents significant challenges, particularly in showcasing the numerical reasoning capabilities of large language models (LLMs) \cite{numerical-reasoning}. When handling queries related to a baby's specific age (For ex- 13 weeks old, 4 months old), both ChatGPT and GPT-4 displayed confusion over the baby's age in generated suggestion questions. Despite the prompt clearly stating that a month consists of 4 weeks, as detailed in Appendix \ref{sec:appendix}, they at times illustrated the difficulty in grasping numerical distinction ( For example- incorrectly interpreted 13 weeks as 13 months).

\section{Prompt Format}
\label{sec:appendix}
This section illustrates the prompt given to the LLMs. 

\textit{You are an intelligent conversational agent for a smart sleep coach of baby. You are given a prompt with a query with different contexts. Each query is one question. Convert the query into question format. You have to suggest 3 different questions that can be very easily answered by only the context provided. If the questions cannot be strictly answered by only the context provided then it should not suggest. Remember in a month there are 4 weeks. The age of the baby should be the same as in the query and not be changed.  Only generate the questions. Here are some examples:} \\

    \{\textit{3 dynamic QAC Dynamic Examples}\} \\

    \{\textit{Query}\} \\

    \{\textit{3 Contexts from which suggestion questions be generated which can be answered from these contexts}.\} \\
  
\section{Ablation Study}
\label{ablation}
\begin{table}[ht]
\centering
    \begin{tabular}{|c|c|c|}
    \hline
           & \textbf{Dynamic Contexts} & \textbf{Order Changed}   \\
         \hline
         GPT4 &46 &46 \\
         \hline
         Claude2 &44 & 44\\
    \hline
    \end{tabular}
    \caption{Ablation Study of Dynamic Context approach after changing the order of the query and the contexts}
    \label{table:Ablation-Study}
\end{table}  

To determine the optimal order of the final prompt, we conducted an ablation study, altering the arrangement of example contexts and queries to assess performance. Initially, we presumed that presenting the query before the dynamic contexts would enhance comprehension, allowing the model to suggest questions based on the given information. However, upon changing the order, we did not observe any discernible difference in the nature of the generated suggestion questions as can be seen from the table \ref{table:Ablation-Study}. 

\section{Preference BenchMarking Results}
Refer Table \ref{table:Preference-BenchMarking}
\label{sec:benchmark}
\begin{table}[ht]
    \begin{tabular}{|p{.08\textwidth}|p{.135\textwidth}|p{.08\textwidth}|p{.1\textwidth}|}
    \hline
           & \textbf{Human Eval} &\textbf{GPT-4} &\textbf{ Claude2 }\\
         \hline
         GPT-4 & 21 (11 both)&21 &24\\
         \hline
         Claude2 & 16 &27 &24\\
    \hline
    \end{tabular}
    \caption{Preference BenchMarking to evaluate the models preference}
    \label{table:Preference-BenchMarking}
\end{table}  

\end{document}